\documentclass{article}

\usepackage[margin=1in]{geometry}
\usepackage[en-US]{datetime2} 
\usepackage[utf8]{inputenc}
\usepackage{graphicx}
\usepackage{amsmath}
\usepackage{hyperref}
\usepackage{tikz}
\usetikzlibrary{positioning}

\raggedright

\title{Domain Adaptation and Reasoning Frameworks in Language Models: A Controlled Experiment with Historical Cosmology}
\author{
Francesco De Bernardis\thanks{Independent Researcher. Correspondence: \texttt{fdeberna@gmail.com}}
}

\begin{document}
\DTMlangsetup{showdayofmonth=false}


\maketitle

\begin{abstract}
We investigate how domain adaptation reshapes explanatory behavior in language models using historical cosmology as a controlled setting. In Phase~1, we train a small language model from scratch on a pre-Copernican corpus from which explicit heliocentric references were removed, and evaluate whether Earth-motion or heliocentric continuations nevertheless emerge. In Phase~2, we fine-tune a larger pretrained model using QLoRA on the same corpus in order to study how adaptation modifies explanatory framing and cosmological stance.

Model outputs are evaluated using an LLM-as-judge framework that labels both cosmological stance (geocentric, heliocentric, or ambiguous) and explanatory frame (premodern versus modern). In the constrained setting of Phase~1, the smaller models occasionally generate local Earth-motion continuations, but these remain globally unstable and insufficient to support coherent cosmological reasoning. In Phase~2, fine-tuning induces a large and statistically significant shift toward premodern explanatory framing, while the conditional cosmological stance distributions remain comparatively stable within those frames. As a result, increases in geocentric outputs arise primarily from redistribution over explanatory regimes rather than from direct modification of stance.

These results suggest that domain adaptation may primarily reshape the linguistic frameworks from which continuations are generated, with changes in stance emerging secondarily from those shifts.

\end{abstract}
\vspace{1em}


\section{Introduction}

There is an ongoing debate over whether language models can generate ideas or explanatory structures that are not directly represented in their training data. One difficulty in studying this question lies in defining what would constitute such a conceptual departure, and how it could be distinguished from memorization, interpolation, or recombination of patterns already present in the training distribution.

Recent work has explored this problem in contexts ranging from scientific ideation to analogical reasoning and hypothesis generation \cite{si2024llmsgeneratenovelresearch,radensky2026humanllmcompoundscientificideation}. However, it remains difficult to determine whether apparently novel outputs reflect genuinely new explanatory organization or merely the retrieval and recombination of latent structures already encoded during pretraining. 

Historical cosmology provides a controlled setting for studying this problem. In many scientific domains, conceptual progress required the introduction of new entities or mechanisms, such as viruses and bacteria replacing the humoral theories of medieval medicine. By contrast, the transition from geocentric to heliocentric astronomy reorganized already existing concepts, such as celestial bodies, planetary motion, and orbital structure, within a different explanatory framework. This makes it possible to ask whether language models trained within constrained historical distributions can nevertheless generate continuations adjacent to conceptual structures absent from the fine-tuning corpus.

More specifically, the present work investigates the relationship between explanatory framing and cosmological stance under constrained domain adaptation. We examine whether fine-tuning primarily changes explicit cosmological commitments, broader explanatory organization, or the interaction between the two.

We conducted an experiment in two phases. In the first phase, we study the behavior of small models trained from scratch under strict data constraints. The models were trained on a historical corpus deliberately restricted to pre-Copernican astronomical material, with explicit heliocentric references removed through filtering and preprocessing. In the second phase, we use a large pretrained model adapted via parameter-efficient fine-tuning on the same astronomy corpus. These complementary settings allow us to probe both local conceptual recombination under constrained training and explanatory-frame selection in large pretrained models.\newline 
In the constrained setting of Phase~1, the smaller models occasionally generated local Earth-motion or heliocentric continuations despite the absence of explicit heliocentric material in the fine-tuning corpus. However, these outputs remained globally unstable and insufficient to support strong claims about sustained conceptual reasoning or coherent cosmological model formation. Most continuations instead consisted of ambiguous or weakly committal premodern astronomical language without a stable cosmological stance.\newline
Across both phases, we found that fine-tuning on a pre-Copernican astronomy corpus did not induce direct shifts toward a geocentric stance. The dominant effect was a redistribution over explanatory and linguistic regimes, with cosmological stance remaining comparatively stable conditional on those regimes. In particular, large pretrained models already contain latent historical astronomical continuations prior to fine-tuning, and domain adaptation altered the probability of entering those explanatory manifolds, with increases in geocentric stance emerging secondarily from the increased sampling of that premodern region.

Taken together, these experiments suggest that explanatory framing and cosmological stance behave as partially separable dimensions of language-model generation, and that domain adaptation can reshape the probability of entering different linguistic explanatory regimes within an already existing conceptual space. Rather than treating generated stance as a single output variable, the present work decomposes generation behavior into explanatory-frame selection and conditional stance expression.

\section{Experimental Design}\label{sec:exp_design}

\subsection{Phase 1: Small-model training}

In Phase 1, we train a 110M-parameter GPT model on a filtered general-language corpus with astronomical content removed, followed by fine-tuning on a smaller pre-Copernican astronomy corpus.

The purpose of the general corpus is to provide broad exposure to English syntax, vocabulary, and discourse structure while minimizing direct exposure to modern astronomical concepts. The corpus was constructed from the Project Gutenberg archive \cite{gutenberg}, selecting a total of 2,851 documents using metadata-based filtering to exclude texts explicitly related to astronomy or science.

Because constructing a sufficiently large training corpus exclusively from Medieval or premodern texts was not feasible using publicly available sources, the general corpus also includes relatively recent literary and historical works. However, metadata filtering alone was insufficient to guarantee the removal of modern astronomical knowledge. For example, otherwise unrelated texts could still contain references to heliocentrism, planetary motion, or modern cosmology.

To reduce this contamination, the selected documents were further processed through keyword and pattern-based filtering designed to remove passages referring to concepts such as Earth's orbit, heliocentrism, Copernicus, Galileo, and related modern astronomical ideas, while preserving general linguistic structure and non-astronomical content from literature, philosophy, and history. While complete removal of all indirect astronomical references cannot be guaranteed, the filtering procedure substantially reduced explicit exposure to heliocentric concepts and modern astronomical explanations.

The astronomy corpus is substantially smaller than the general corpus due to the limited availability of publicly accessible English translations of pre-Copernican texts. The corpus combines classical, late antique, and Medieval works containing geocentric astronomical reasoning, cosmological discussion, and premodern natural philosophy. Representative texts include Sacrobosco’s \textit{Sphaera Mundi}, Ptolemy’s \textit{Almagest}, Plato’s \textit{Timaeus}, Aristotle’s \textit{De Caelo}, Cleomedes’s \textit{On the Heavens}, and Peuerbach’s \textit{Theoricae Novae Planetarum}. Some works are explicitly astronomical, while others embed geocentric cosmology within broader philosophical or theological discussions. For a more comprehensive list see Appendix~\ref{appendix:astronomy_corpus}. Because many modern translations include editorial notes, annotations, or footnotes containing references to contemporary astronomy, the same pattern-based filtering procedure was applied to the astronomy corpus in order to remove translator commentary and other potential sources of modern astronomical leakage.

The cleaned corpus was tokenized using a byte-level BPE tokenizer with a vocabulary size of 32,000, trained on the combined general and astronomy corpora. Four special tokens were added: \texttt{<doc>}, \texttt{<pad>}, \texttt{<bos>}, and \texttt{<eos>}. Documents were tokenized as plain text and cached into fixed-length streams for subsequent pretraining and fine-tuning.

We first trained the model on the filtered general corpus only. This process yielded what we refer to as \textit{Model A}. The model is a decoder-only GPT-style transformer with 12 layers, 12 attention heads, embedding dimension 768, and context length 1024, corresponding to approximately $110$M parameters.

\textit{Model B} was initialized from the final general-pretraining checkpoint and fine-tuned on a mixture of pre-Copernican astronomy text and general English text. Each fine-tuning batch was sampled with an astronomy/general mixture ratio of 0.8/0.2, allowing the model to adapt to the astronomy corpus while retaining general language fluency. Fine-tuning used the same tokenizer, context length, and model architecture as pretraining. Training was performed for 20,000 iterations using AdamW with learning rate $5\times10^{-5}$, cosine decay to $10^{-6}$, 500 warmup iterations, batch size 12, and gradient accumulation over 4 steps.
 

\subsection{Phase 2: QLoRA adaptation}

In Phase~2, we adapt Qwen2.5-7B \cite{qwen25} using QLoRA \cite{dettmers2023qlora} on the same pre-Copernican astronomy corpus. Qwen2.5-7B is a modern decoder-only large language model pretrained on a broad multilingual corpus and exhibits stronger linguistic fluency and reasoning stability than the $110$M models used in Phase~1. 

Using a large pretrained model introduces an important trade-off. While the pretrained model provides a robust and highly stable language prior, it also likely contains latent representations associated with modern astronomical concepts acquired during large-scale pretraining. Fine-tuning on the pre-Copernican corpus may suppress or overwrite some of these representations, but cannot guarantee their complete removal. This limitation should therefore be considered when interpreting generated outputs. 

The base Qwen model was loaded in 4-bit NF4 quantization with double quantization and trained using LoRA adapters. We used rank $r=16$, LoRA scaling $\alpha=32$, dropout $0.05$, and no bias adaptation. Adapters were applied to both attention projections (\texttt{q\_proj}, \texttt{k\_proj}, \texttt{v\_proj}, \texttt{o\_proj}) and MLP projections (\texttt{gate\_proj}, \texttt{up\_proj}, \texttt{down\_proj}). Text files were split by document into training and validation sets, with a validation ratio of $0.05$, then tokenized and chunked into causal language-modeling sequences of length 512. We trained two adapters, one for 500 steps and one for 1000 steps, using per-device batch size 1, gradient accumulation over 16 steps, learning rate $2\times10^{-4}$, warmup ratio $0.03$, fp16 precision, gradient checkpointing, max gradient norm $0.3$, and the paged AdamW 8-bit optimizer. We will refer to these models as \textit{QLoRA-500} and \textit{QLoRA-1000} respectively.

\subsection{LLM-as-judge}
Across Phase~1 and Phase~2, we evaluate five models: Model~A and Model~B from Phase~1, together with the Base Qwen, QLoRA-500, and QLoRA-1000 models from Phase~2.

Generated outputs were evaluated using an LLM-as-a-judge framework, in which a separate language model was prompted to assign structured labels describing cosmological stance, explanatory frame, Earth-motion references, and ambiguity. This approach follows recent work \cite{huang-etal-2025-empirical,zheng2023judging} showing that large language models can provide useful and scalable comparative evaluations for open-ended generation tasks. All generations were evaluated using the Claude Haiku 4.5 model (\texttt{claude-haiku-4-5-20251001}) as an LLM judge.

Because small-model generations were often truncated, contradictory, or partially incoherent, the evaluation rubric was designed to detect local semantic signals rather than requiring globally coherent reasoning.

The judge model assigned structured labels capturing stance, explanatory framing, and Earth-motion references. The primary labels used in this work are summarized below:

\begin{itemize}
    \item \textbf{quality score}: 
    A coarse coherence score from 0 to 2, where 0 indicates degenerate or uninterpretable output, 1 indicates partially coherent but confused or contradictory astronomy-related text, and 2 indicates coherent and prompt-relevant continuation.

    \item \textbf{earth-motion mention}: 
    A low-threshold lexical signal indicating that the generated continuation explicitly mentions Earth moving, rotating, revolving, or otherwise possessing motion, regardless of whether the surrounding reasoning is coherent.

    \item \textbf{explicit Earth-motion}: 
    A higher-confidence label requiring a locally unambiguous assertion that Earth moves or rotates, rather than merely posing the possibility hypothetically or rhetorically.

    \item \textbf{proto-heliocentric}: 
    Assigned when the text substantively develops or entertains Earth motion or a Sun-centered interpretation as an explanatory possibility, even if the reasoning is incomplete or internally inconsistent.

    \item \textbf{geocentric}: 
    Assigned when the text clearly favors a stationary or Earth-centered cosmology while not triggering Earth-motion labels.

    \item \textbf{ambiguous}: 
    Assigned when astronomical content is present but no stable cosmological stance can be confidently inferred.
\end{itemize}

Additional consistency constraints were enforced automatically between labels (e.g., explicit Earth-motion implies Earth-motion mention and proto-heliocentric stance). Full evaluation prompts, judging scripts, and analysis code are available in the project repository:
\url{https://github.com/fdeberna/chat-ptolemaic}.

Phase~2 introduced additional evaluation labels intended to separate cosmological stance from explanatory framework and stylistic register. During preliminary analysis, it became clear that large pretrained models could generate strongly premodern astronomical language without committing to a stable geocentric stance, or alternatively combine incompatible modern and premodern explanatory systems within the same continuation. The refined labels used in Phase~2 include:

\begin{itemize}

\item \textbf{refined geocentric stance}: 
assigned when the continuation clearly supports an Earth-centered or Earth-stationary cosmology.

\item \textbf{refined heliocentric stance}: 
assigned when the continuation clearly supports Earth motion or a heliocentric ordering.

\item \textbf{refined ambiguous stance}: 
assigned when astronomy-related content is present, but no stable cosmological position can be confidently inferred.

\item \textbf{premodern explanatory frame}:
assigned when the completion explains astronomical phenomena using explicitly premodern cosmological or astronomical machinery, such as celestial spheres, epicycles, deferents, firmaments, crystalline heavens, or historically geocentric explanatory structures.

\item \textbf{modern explanatory frame}: 
assigned when the completion explains astronomical phenomena using modern heliocentric or orbital mechanics concepts, such as planets orbiting the Sun or relative orbital motion.

\end{itemize}

This refined evaluation schema becomes necessary in Phase~2 because the pretrained Qwen model already contains strong modern astronomical and linguistic priors. Unlike the constrained Phase~1 models, which were trained almost entirely within a premodern astronomical distribution, the Phase~2 model can dynamically switch between modern and premodern explanatory regimes, combine incompatible frameworks, or adopt premodern stylistic forms without committing to a stable geocentric stance. Separating explanatory frame, stylistic register, and cosmological stance therefore becomes essential for interpreting the generated outputs. The refined evaluation schema was designed specifically to distinguish distributional shifts in explanatory language from shifts in cosmological stance. 

Note that the refined labels are not designed to be mutually exclusive. Additionally, although the premodern explanatory frame is defined using historical astronomical vocabulary and explanatory structures commonly associated with geocentric cosmology, it is not defined in terms of cosmological stance itself. A continuation may therefore receive a strong premodern label while remaining stance-ambiguous or even partially heliocentric. As we will show, premodern framing and geocentric stance are positively correlated but not equivalent. The same holds, symmetrically, for modern explanatory framing and heliocentric stance.

Phase~1 models were evaluated using four categories of prompts. Each category contains 28 distinct prompts, and 15 generations were sampled per prompt, yielding a total of $4 \times 28 \times 15 = 1{,}680$ responses per model. The four categories are \textit{astronomy}, \textit{declarative}, \textit{general}, and \textit{questions}.

Prompts were formulated as sentence continuations in which the model was asked to complete a partially written statement. The prompt categories were designed to probe different modes of generation and activation. Astronomy-focused prompts directly invoke cosmological reasoning, whereas direct questions may encourage explicit stance-taking. More general or declarative prompts, by contrast, test whether geocentric or heliocentric patterns emerge in less constrained linguistic settings. Prompt categories were evaluated independently in addition to aggregate analysis across all prompts.

Prompts were intentionally written in a premodern or scholastic linguistic style in order to remain consistent with the historical training corpus while avoiding explicit heliocentric framing. Many astronomy prompts were intentionally phrased around historically shared observational phenomena (e.g., retrograde planetary motion) rather than explicitly geocentric doctrine, allowing multiple explanatory regimes to potentially emerge. For example, the astronomy prompts include:

\begin{quote}
\small
``When the wandering stars appear at times to go backward against the firmament, ...''

``Concerning the seeming retrogression of the planets, it must be said that ...''

``When Mars, Jupiter, or Saturn seem to reverse their course for a time, ...''

``The planets are sometimes seen to move contrary to their former progress, and this is because ...''
\end{quote}

For the evaluation of the Qwen models of Phase~2, we used the same 4 categories and 28 prompts, generating 5 samples per prompt, for a total of $4 \times 28 \times 5 = 560$ prompts for each of the 3 models (Base, QLoRA-500, QLoRA-1000).

Generation settings differed slightly between Phase~1 and Phase~2 due to the substantially different stability and coherence properties of the underlying models. The smaller $110$M models in Phase~1 required stronger repetition controls and slightly higher sampling temperature in order to maintain diversity while avoiding degenerate looping behavior. By contrast, the pretrained Qwen2.5-7B model produced more stable continuations, allowing lower temperature and weaker repetition penalties. 

Phase~1 generations used temperature $0.7$, top-$p$ sampling with $p=0.9$, repetition penalty $1.15$, no-repeat trigram constraints, and maximum continuation length of 100 tokens. 

Phase~2 generations used temperature $0.6$, top-$p$ sampling with $p=0.9$, repetition penalty $1.1$, no-repeat 4-gram constraints, and maximum continuation length of 150 tokens.

To evaluate the sensitivity of the results to annotation definitions, in Phase~2, we additionally tested a stricter variant of the judge prompt in which premodern explanatory framing required explicit historical explanatory machinery rather than stylistic or lexical cues alone, and geocentric stance required clearer Earth-centered cosmological commitment.

\section{Results}

\subsection{Phase 1: Controlled Small-Model Experiments}
We begin by analyzing the outputs of the small Phase~1 models trained under strict corpus constraints. We find that both the general English pre-trained model A and its astronomy fine-tuned version (model B), generate heliocentric-like content at low rates ($4\%$--$8\%$), suggesting some pattern recombination occurs even in a small 110M parameter model trained on geocentric texts. Under the assumption that heliocentric content was successfully removed from the training corpus, the emergence of this content must arise through recombination of patterns learned from the remaining corpus. However, the quality of the generated output remains mediocre, with the overwhelming majority ($\sim 94\%$) of the output being labeled with the intermediate score of $1$, representing 
only "partly coherent" output according to our judge manifest. This is not surprising and a known limit of small models (see for example \cite{eldan2023tinystoriessmalllanguagemodels}).

Representative local continuations from Phase~1 are shown below. Although the generations are often globally unstable or partially incoherent, they nevertheless occasionally produce locally meaningful astronomical structure associated with both geocentric and Earth-motion reasoning.

\paragraph{Example 1 - Model B Output.}
\begin{quote}
\textbf{Prompt:} \\
\textit{The planets are sometimes seen to move contrary to their former progress, and this is because}

\vspace{0.5em}

\textbf{Completion:} \\
\textit{The planets are sometimes seen to move contrary to their former progress, and this is because the center of the epicycle is now very near the mean distances of the deferent, so that when they move in a straight line, namely, one degree to the east or west, then other degrees will occur. The equation of longitude. The situation of Saturn. When indeed he is on the meridian above the earth, as is seen above the deceptive circle, his superior planets are observed to be north of the ecliptic by little and little. This is due to their motion from the}

\vspace{0.5em}

\textbf{Judge label:} geocentric.
\end{quote}

\paragraph{Example 2 - Model B Output.}
\begin{quote}
\textbf{Prompt:} \\
\textit{The fact that the seasons return in fixed order is connected with the solar course, for}

\vspace{0.5em}

\textbf{Completion:} \\
\textit{The fact that the seasons return in fixed order is connected with the solar course, for Saturn retires to the north when he is in Leo, but for Jupiter to the south when he recedes from it. Therefore the obliquity of the zodiac relative to the ecliptic. But because the whole earth moves on its axis, and revolves naturally about one center, therefore this motion appears to be that which we call a fixed sphere. And since indeed no other natural body has any regular movement beyond the heaven, as is said above (vertically) we also may consider all those movements}

\vspace{0.5em}

\textbf{Judge label:} heliocentric.
\end{quote}

The first continuation locally reproduces clear Ptolemaic astronomical vocabulary, including epicycles, deferents, and ecliptic motion, despite broader instability in the continuation. The second shows a clear assertion of Earth's motion and revolution despite incoherence in the rest of the sentence.

The more interesting result is that Model A, despite being trained on an astronomy-free corpus, shows higher heliocentric content than Model B. Specifically, higher Earth-motion mentions rates ($8.3\%$ for model A vs $5.7\%$ for model B) and higher explicit Earth-motion mentions ($4.2\%$ (A) vs $3.3\%$ (B))
and proto-heliocentric content ($5.1\%$ (A) vs $4.0\%$ (B)). At the same time, surprisingly, model B also shows a lower rate of geocentric content ($3.3\%$ for A vs $1.9\%$ for B). This apparent contradiction is resolved by the observation that Model B produces substantially more ambiguous and hedged outputs than Model A. The fraction of \textit{Ambiguous} labels increased from $41\%$ (Model A) to $54\%$ (Model B). Therefore, fine-tuning introduced more ambiguity instead of reinforcing geocentric doctrine. Model B did not shift towards either a heliocentric or geocentric perspective.

In order to assess the statistical significance of these results, we conducted a prompt-level paired permutation test \cite{good2005permutation}. Since each prompt was sampled $15$ times per model, the 1680 generations per model are not independent observations. Some prompts are intrinsically more likely than others to elicit Earth-motion or ambiguity, and those within-prompt samples are correlated. For each of the $112$ prompts, the average rate $r$ for a given label was computed separately for A and B. This yields one paired difference per prompt, for each label $\ell$:

$$d^{\ell}_{AB} = r^{\ell}_{A} - r^{\ell}_{B}$$

The goal of the permutation test is to compare the observed mean paired difference with the distribution expected under the null hypothesis that Models A and B are exchangeable at the prompt level. To accomplish this, the test flipped the sign of each paired difference $10000$ times, at random, and recalculated the mean. The two-sided $p$-value is the proportion of random sign-flips whose absolute mean difference is at least as large as the observed one. The advantage of this method is that it is a non-parametric test, requiring no assumption about the distribution of the differences. Table \ref{tab:phase1_results} shows the most relevant values. The drop in both \textit{Earth-motion} mentions and \textit{Geocentric} labels is significant, but the most statistically significant result is the increase in \textit{Ambiguous} statements. For a detailed analysis by prompt category see Table~\ref{tab:phase1_category_results} in Appendix~\ref{appendix:prompt_analysis}.

\begin{table}[t]
\centering
\setlength{\tabcolsep}{4pt}
\caption{Comparison of heliocentric and ambiguity-related behaviors between Model A (general pre-training only) and Model B (astronomy fine-tuned). Values are averaged over paired prompts.}
\label{tab:phase1_results}
\begin{tabular}{lcccc}
\hline
Label & A & B & $\Delta_{BA}$ & $p$ \\
\hline
Earth-motion          & 8.3\%  & 5.7\%    & -2.6\%  & 0.0226$^{**}$ \\
Explicit Earth-motion & 4.2\%  & 3.3\%    & -0.9\%  & 0.0954 \\
Proto-heliocentric    & 5.1\%  & 4.0\%    & -1.1\%  & 0.1500 \\
Geocentric            & 3.3\%  & 1.9\%    & -1.4\%  & 0.0136$^{**}$ \\
Ambiguous             & 41.0\% & 54.8\%   & +13.8\% & $<10^{-4\,***}$ \\
\hline
\end{tabular}
\vspace{0.5em}
\footnotesize{
$^{**} p < 0.05$, \quad
$^{***} p < 0.001$
}
\end{table}

\begin{figure}[t]
\centering
\includegraphics[width=\columnwidth]{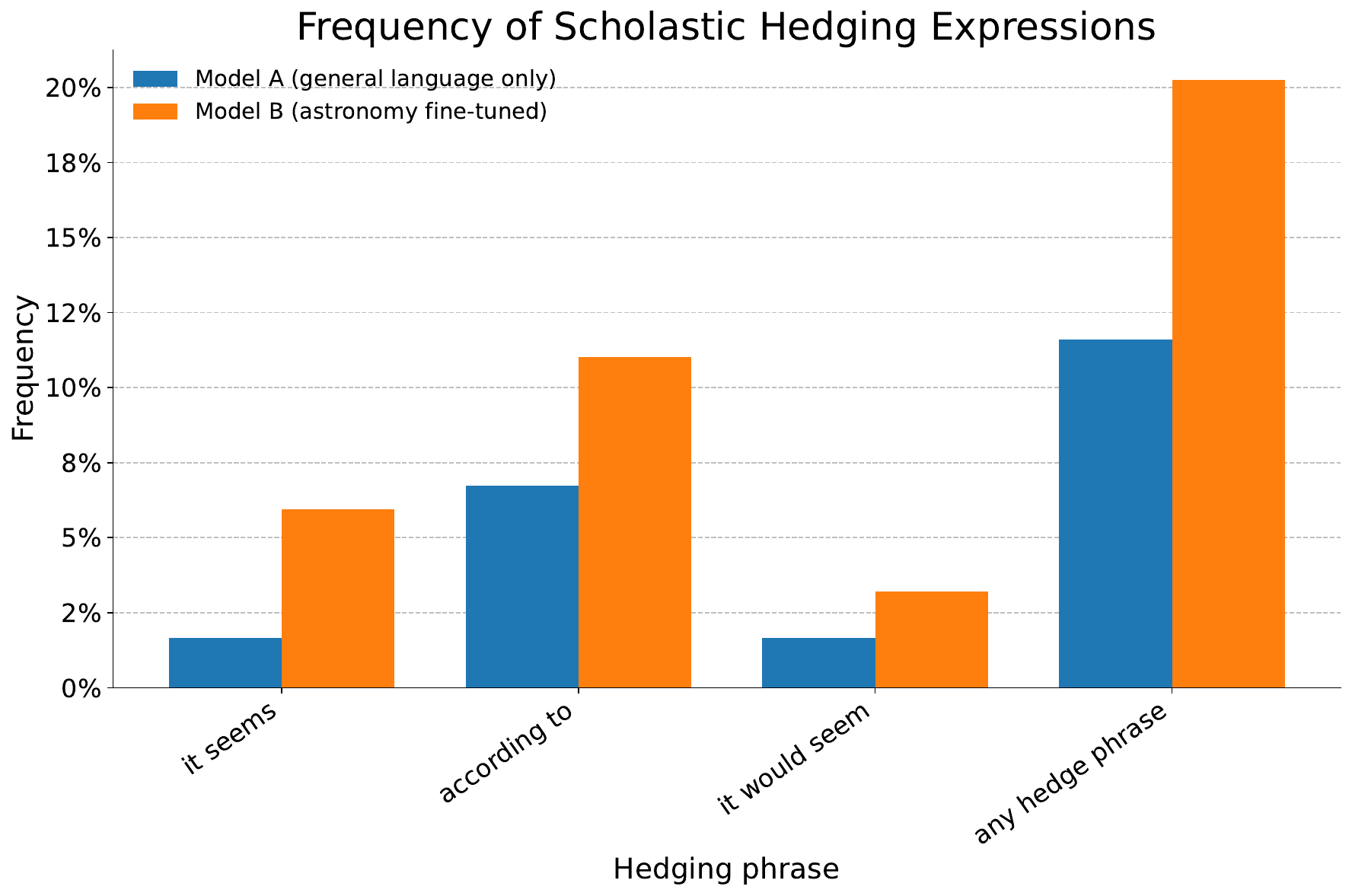}
\caption{Frequency of selected scholastic hedging expressions in generated outputs from Model A (general pre-training only) and Model B (astronomy fine-tuned). Astronomy fine-tuning substantially increases the use of non-committal and interpretive language patterns characteristic of premodern scholarly discourse.
}
\label{fig:hedging_phrases}
\end{figure}

The increased ambiguity of Model B can be further confirmed by a comparison of the frequency of non-committal, qualified statements and expressions between Model A and B, as shown in Figure~\ref{fig:hedging_phrases}. The rate of any hedge phrase increased from $11.6\%$ in Model A to $20.2\%$ in B. This increase appeared across all categories. In particular, phrases like \textit{it seems}, \textit{according to}, and \textit{it would seem} became noticeably more frequent in B. This supports the interpretation that fine-tuning on geocentric astronomy content increased qualified, scholastic, stance-uncertain astronomy discourse, but did not strengthen a stable geocentric doctrine. The primary effect of astronomy fine-tuning was a shift toward more qualified and interpretive explanatory discourse.

\subsection{Phase 2: QLoRA Adaptation of Large Models}

We use the same prompts as in Phase~1 to compare three settings:

\begin{itemize}
    \item Qwen2.5-7B base model;
    \item Qwen2.5-7B fine-tuned for 500 iterations of QLoRA on the pre-Copernican corpus (QLoRA-500);
    \item Qwen2.5-7B fine-tuned for 1000 iterations of QLoRA on the pre-Copernican corpus (QLoRA-1000).
\end{itemize}

In contrast to Phase~1, where models were trained from scratch under strict data constraints, Phase~2 investigates how large pre-trained models behave when exposed to the same domain-specific fine-tuning. The generated continuations are substantially more coherent than those observed in Phase~1, with $>95\%$ having at least a quality score of $1$ and $\sim35\%$ of the responses having a score of $2$, which was never reached in Phase~1.

A first striking result is that the base model already produces approximately $8\%$ of outputs classified as \textit{geocentric}. In several cases, these outputs are not merely consistent with a geocentric framework, but explicitly instantiate classical geocentric cosmology, including the motion of celestial spheres around the Earth. These outputs typically arise under prompts that strongly activate historical or scholastic explanatory regimes, suggesting that such frameworks are already present in the pretrained model and can be selectively activated. A typical example is provided below.

\paragraph{Example: Base Qwen model, no fine-tuning.}
\begin{quote}
\textbf{Prompt:} \\
\textit{Of the celestial motions and the place of the Earth, one may begin thus, that}

\vspace{0.5em}

\textbf{Completion:} \\
\textit{the common people have no knowledge of the matter, but only by hearsay are they told that there is a great sphere in heaven above us which revolves round the earth; for this is the popular opinion. They think that it is like a wheel or rather a plate, and that it has an axle at its centre, upon which it turns and carries with it all the stars. Astronomy, the science of the heavens, is divided into two parts, practical and theoretical. The former is concerned with the use of the instruments of observation, the latter with the actual phenomena observed, and with their causes. We must therefore treat first of the nature of the heavenly bodies, secondly of the instruments used to observe them, thirdly of the places}
\vspace{0.5em}

\textbf{Judge labels:} geocentric, premodern.
\end{quote}

This example shows how the generated output matches the pre-modern register of the prompt, with a geocentric cosmology involving spheres revolving around the Earth.

These geocentric outputs of the base model are mostly caused by astronomical and declarative prompts. The geocentric rate of the base model for each mode category is the following:
\begin{itemize}
    \item Astronomy prompts: $13\%$;
    \item Declarative prompts: $15\%$;
    \item Questions: $5\%$;
    \item General prompts: $0\%$.
\end{itemize}

These observations suggest that the base model already contains latent historical explanatory modes that can be activated by appropriately premodern prompts, even prior to fine-tuning. This is an important consideration for interpreting the fine-tuned outputs, because strongly premodern astronomical language does not uniquely determine cosmological stance in Phase~2 generations. As the following examples illustrate, premodern explanatory structure may coexist with ambiguous, geocentric, hybrid, or explicitly Earth-motion reasoning. Distinguishing explanatory framing from explicit cosmological stance, therefore, becomes necessary for interpreting the behavior of the fine-tuned model.

\paragraph{Example: QLoRA-1000.}
\begin{quote}
\textbf{Prompt:} \\
\textit{The Sun is seen to pass through the zodiac in a yearly circuit, and the cause of this is}

\vspace{0.5em}

\textbf{Completion:} \\
\textit{due to the motion of the earth. For as it turns round its axis, which lies from east to west, so that one half of it may always be turned towards the sun, while the other is turned away from it, there are two opposite points on the equator (the line drawn across the middle of the earth) where the day lasts for twelve hours: and these points are called the solstices. And when the centre of the Sun is at either of those points, the whole earth is divided into night and day by the meridian line, which runs through both poles of the world; but if the centre of our orb were not moved about its own axis, the Sun would only appear once every month or year. But since}
\vspace{0.5em}

\textbf{Judge labels:} heliocentric, premodern.
\end{quote}

\paragraph{Example: QLoRA-1000.}
\begin{quote}
\textbf{Prompt:} \\
\textit{Since the wandering stars hasten and delay in unequal manner, it is necessary to say that}

\vspace{0.5em}

\textbf{Completion:} \\
\textit{they are not carried by their own motion. Therefore they are moved by another; but it seems impossible for them to be moved by any other than the Sun. For if we suppose that the Moon were moved by Mars, or Jupiter by Venus, or Saturn by Mercury, then one of these would have to precede its proper order, as when a man runs faster than his companion. But this does not happen: for the planet which precedes has always been found to be slower than the one following. It must therefore be granted that all the wandering stars are moved by the same power. It also appears that the planets cannot be moved either by the same force or in the same way as the fixed stars, since the latter move with uniform speed on}
\vspace{0.5em}

\textbf{Judge labels:} ambiguous, premodern.
\end{quote}

The examples above illustrate that even when a premodern explanatory regime is activated, the stance distribution within that regime exhibits nontrivial structure. The first example shows a period-style prose while maintaining a clear heliocentric stance. The second example remains ambiguous, displaying a coherent premodern astronomical discussion about planetary motion without asserting Earth's motion nor developing a clear heliocentric stance, despite suggesting the Sun as a possible organizing principle for planetary motion, which does not by itself imply a heliocentric cosmology.

These examples motivate the conditional analyses presented below, where explanatory register and cosmological stance are treated as partially separable dimensions of generation behavior.

The main finding of Phase~2 is that the QLoRA fine-tuning produces a shift toward premodern explanatory discourse, accompanied by a sharp suppression of heliocentric reasoning and only a modest increase in geocentric outputs (Table \ref{tab:phase2_summary}).

\begin{table*}[t]
\centering
\caption{Phase 2: Effect of QLoRA fine-tuning on explanatory frame and stance. The reported \textit{Geocentric} or \textit{Heliocentric} stances are the refined stricter version of the Phase~1 labels.}
\label{tab:phase2_summary}

\begin{tabular}{lccc}
\hline
Model & Geocentric & Heliocentric & Premodern \\
\hline
Base & 8.2\% & 21.4\% & 35.2\% \\
QLoRA-500 & 14.6\% & 5.0\% & 64.5\% \\
QLoRA-1000 & 15.9\% & 2.7\% & 65.4\% \\
\hline
\end{tabular}
\end{table*}

The observed rates show how the main effect of fine-tuning is to shift the explanatory framework towards premodern language, with the change in stance being only secondary in magnitude. Additionally, we observe that there is minimal difference between QLoRA-500 and QLoRA-1000, with most of the effects of fine-tuning being visible after 500 iterations only.

\begin{table*}[h]
\centering
\caption{Relationship between explanatory frame and stance.}
\label{tab:frame_stance_conditionals}
\begin{tabular}{lcccc}
\hline
Model & $P(\mathrm{Premodern})$ & $P(\mathrm{Geo}\mid\mathrm{Pre})$ & $P(\mathrm{Helio}\mid\mathrm{Pre})$ & $P(\mathrm{Helio}\mid\mathrm{Modern})$ \\
\hline
Base & 30.9\% & 24.3\% & 12.7\% & 68.7\% \\
QLoRA-500 & 64.5\% & 22.7\% & 3.9\% & 71.0\% \\
QLoRA-1000 & 65.4\% & 23.5\% & 2.2\% & 66.7\% \\
\hline
\end{tabular}
\end{table*}

In our analysis of the impact of QLoRA across various prompt categories, we observed significant variability in its effects. The increase in geocentric outputs is strongest for astronomy prompts, while general prompts remain largely unaffected. A detailed breakdown by category is provided in Table~\ref{appendix:phase2_categories} in Appendix~\ref{appendix:prompt_analysis}. Therefore, the choice of prompt type determines how visible the effects of the fine-tuning are. Prompts that already encourage astronomical reasoning exhibit the largest increase in geocentric outputs, while general prompts show minimal change in stance. However, the shift in explanatory register is strong across all the prompt categories, regardless of the cosmological stance.

The increase in geocentric outputs and the decrease in heliocentric outputs are modest in comparison to the induced change in explanatory frame. Given the strong shift toward premodern explanatory language under QLoRA, it is useful to decompose the probability of a geocentric output as follows.

\begin{equation*}
\begin{aligned}
P(\mathrm{Geo})
&= P(\mathrm{Geo}\mid \mathrm{Premodern})\,P(\mathrm{Premodern}) \\
&\quad + P(\mathrm{Geo}\mid \neg \mathrm{Premodern})\,P(\neg \mathrm{Premodern}) .
\end{aligned}
\end{equation*}

We find that $P(\mathrm{Geo}\mid \neg \mathrm{Premodern})$ is negligible for all models (on the order of $0$–$1\%$), so that we can approximate:

\begin{equation}\label{eq:geo_decomposition_approx}
P(\mathrm{Geo}) \simeq P(\mathrm{Geo}\mid \mathrm{Premodern})\,P(\mathrm{Premodern})
\end{equation}

Additionally, the conditional probabilities of stance given explanatory frame remain largely stable across models (see Table~\ref{tab:frame_stance_conditionals}). We emphasize that, as demonstrated by the examples at the beginning of this section, the premodern explanatory frame, while defined using historical astronomical machinery associated with geocentric models, does not imply a geocentric stance (see also Section~\ref{sec:exp_design}). Most premodern-frame outputs remain ambiguous or non-committal, allowing the relationship between frame and stance to be evaluated empirically rather than being fixed by definition.

In particular, $P(\mathrm{Geo}\mid\mathrm{Premodern})$ remains stable across models, and $P(\mathrm{Helio}\mid\mathrm{Modern})$ varies only marginally. In contrast, $P(\mathrm{Helio}\mid\mathrm{Premodern})$ decreases sharply under QLoRA. This reduction is primarily absorbed by an increase in ambiguous or non-committal outputs within the premodern frame, rather than by a corresponding increase in geocentric assertions.

\begin{figure*}[hbt]
\centering
\begin{tikzpicture}[
    node distance=3.5cm,
    box/.style={draw, rounded corners, align=center, minimum width=3.5cm, minimum height=1.2cm},
    arrow/.style={->, thick}
]

\node[box] (prompt) {Prompt};
\node[box, right=of prompt] (frame) {Explanatory\\Frame};
\node[box, right=of frame] (stance) {Cosmological\\Stance};

\draw[arrow] (prompt) -- node[above] {$P(f \mid x, m)$} (frame);
\draw[arrow] (frame) -- node[above] {$P(s \mid f, x, m)$} (stance);

\node[align=center, below=1.2cm of frame] (qlora) {QLoRA effect};

\draw[arrow, dashed] (qlora) -- (frame);

\end{tikzpicture}
\caption{Conceptual decomposition of model behavior into explanatory frame selection and stance realization. Given a prompt $x$ and model $m$, the model first induces a distribution over explanatory frames $f \in \{\mathrm{Premodern}, \mathrm{Modern}\}$ via $P(f \mid x, m)$, and then produces a cosmological stance $s \in \{\mathrm{Geocentric}, \mathrm{Heliocentric}, \mathrm{Ambiguous}\}$ according to $P(s \mid f, x, m)$. Our results indicate that QLoRA primarily shifts the distribution over frames $P(f \mid x, m)$, increasing the probability of premodern framing, while leaving the conditional mapping from frame to stance largely unchanged.}
\label{fig:frame_stance_model}
\end{figure*}
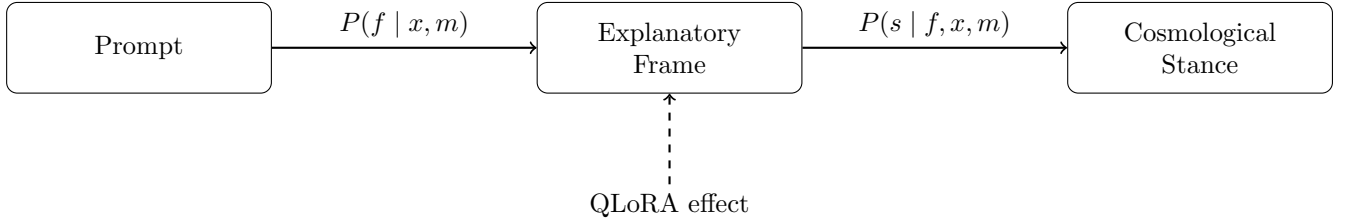

\begin{figure*}[h!]
\centering
\includegraphics[width=\textwidth]{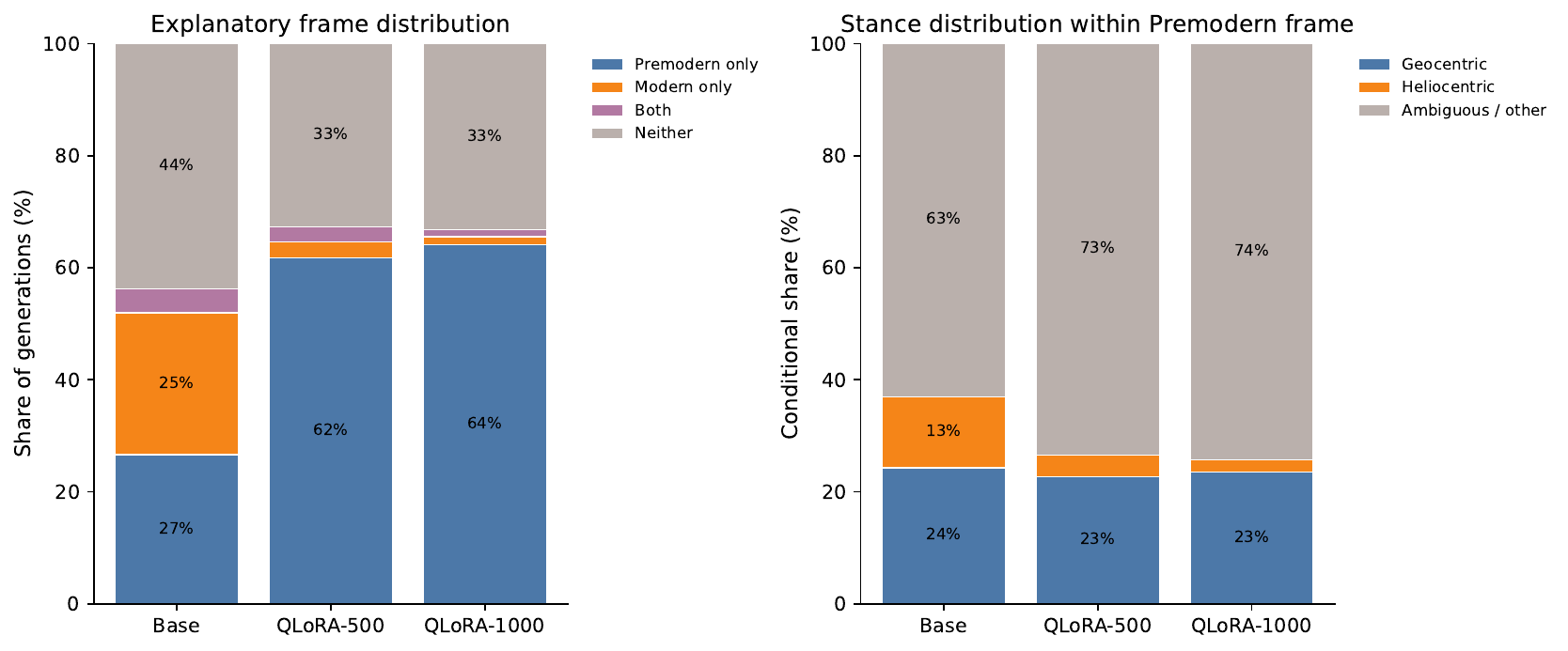}
\caption{
Distribution over explanatory frames (left) and conditional stance distribution within the premodern frame (right). QLoRA substantially increases the probability of premodern framing, while the conditional distribution over stances within that frame remains comparatively stable. The dominant effect of fine-tuning is therefore a redistribution over explanatory regimes rather than a direct modification of stance within regimes.
}
\label{fig:stacked_bars_labels}
\end{figure*}

Combined with the substantial increase in premodern language (Table~\ref{tab:phase2_summary}), these results indicate that fine-tuning does not significantly modify the stance within the same explanatory frame. Instead, it shifts the distribution over explanatory regimes, substantially increasing the probability $P(\mathrm{Premodern})$ of entering a premodern explanatory frame, while leaving the conditional mapping from frame to stance unchanged. As a result, the overall probability of geocentric outputs $P(\mathrm{Geo})$ in (\ref{eq:geo_decomposition_approx}) increases primarily through this redistribution, despite largely stable conditional structure within frames. This process is summarized in Figure~\ref{fig:frame_stance_model}. The distribution of explanatory frames between different models and of stance labels conditional to the premodern frame is summarized in Figure~\ref{fig:stacked_bars_labels}

We can further support these conclusions by examining paired transition rates between models, i.e., how often the response to the same prompt changes explanatory frame or stance between the base model and the QLoRA fine-tuned models. Because each comparison uses generations produced from identical prompts, these transition rates isolate the effect of fine-tuning conditional on a fixed input distribution. Statistical significance is assessed using McNemar’s test \cite{mcnemar1947}, which is appropriate for paired categorical responses and evaluates whether transitions in one direction occur significantly more often than transitions in the reverse direction. The evaluation consists of 28 prompts for each of 4 prompt categories, with 5 sampled generations per prompt, yielding a total of 28×4×5=560 paired generations for each model comparison. These results are summarized in Table~\ref{tab:flip_rates}.

For the same prompt, we observe that the directional transition from modern to premodern explanatory framing occurs in approximately $25\%$ of paired generations and is highly statistically significant. The reverse transition is rare ($<2\%$). In contrast, the transition from \textit{heliocentric} to \textit{geocentric} stance, while also statistically significant, occurs at a substantially lower rate. Similarly, earth-motion suppression is strongly significant, indicating that QLoRA reduces explicit heliocentric reasoning and shifts many responses toward ambiguous, non-committal, or geocentric interpretations.

The transition rates reported in Table~\ref{tab:flip_rates} correspond to the comparison between the base model and QLoRA-1000. However, when comparing paired responses between the base model and QLoRA-500, we observe qualitatively similar transition asymmetries with comparable statistical significance ($p < 10^{-3}$), indicating that the explanatory frame and stance shifts emerge within a relatively small number of fine-tuning iterations.

\begin{table*}[t]
\centering
\caption{Pairwise transition rates between the base model and QLoRA-1000. Reported rates correspond to the fraction of paired generations exhibiting the specified transition. p-values are computed using McNemar's test.}
\label{tab:flip_rates}

\begin{tabular}{lcccc}
\hline
Transition & Count & Rate & Reverse Count & McNemar $p$ \\
\hline

Modern $\rightarrow$ Premodern
& 141/560
& 25.2\%
& 9
& $<10^{-6}$ \\

Premodern $\rightarrow$ Modern
& 9/560
& 1.6\%
& 141
& $<10^{-6}$ \\

Heliocentric $\rightarrow$ Geocentric
& 19/560
& 3.4\%
& 2
& $4.8\times10^{-4}$ \\

Geocentric $\rightarrow$ Heliocentric
& 2/560
& 0.4\%
& 19
& $4.8\times10^{-4}$ \\

Earth-motion suppression
& 100/560
& 17.9\%
& 30
& $<10^{-6}$ \\

Earth-motion activation
& 30/560
& 5.4\%
& 100
& $<10^{-6}$ \\

\hline
\end{tabular}
\end{table*}

\subsection{Robustness to Alternative Label Definitions}


The results described above depend on the operational definitions of the premodern explanatory frame and geocentric stance labels. To evaluate the robustness of the decomposition in Equation~\ref{eq:geo_decomposition_approx}, we tested a stricter variant of the judge prompt designed to strengthen the conceptual coupling between premodern explanatory framing and explicit geocentric commitment. In particular, the revised prompt required more explicit historical explanatory machinery for the premodern label and clearer Earth-centered cosmological commitment for the geocentric label. The modified definitions are shown in Appendix~\ref{label:alt-prompts}.

Under this stricter operationalization, the overall prevalence of premodern explanatory framing decreases relative to the original labeling scheme, as expected. However, the central qualitative pattern remains unchanged. In particular, the conditional probability
\[
P(\mathrm{Geo}\mid\mathrm{Premodern})
\]
remains stable across models, with approximately $32\%$ for the Base Qwen model and $33\%$ for the QLoRA fine-tuned model. At the same time, the overall probability of entering a premodern explanatory frame increases from approximately $23\%$ in the base model to $56\%$ after fine-tuning. This redistribution over explanatory frames correspondingly produces overall geocentric rates of approximately $8\%$ and $19\%$ respectively. The robustness analysis therefore supports the same qualitative interpretation observed under the original labeling scheme: fine-tuning primarily reshapes the probability of entering particular explanatory frameworks, while the conditional relationship between explanatory frame and cosmological stance remains comparatively stable.
\section{Discussion}

The findings of Phase~1 and Phase~2 are complementary. Phase~1 focused on a comparatively small $110$M-parameter language model, enabling tighter control over the training corpus and the astronomical knowledge available to the model, at the cost of the well-known limitations of small-scale language models and substantially less stable generations.

Within this constrained setting, the model occasionally produced local recombinations adjacent to heliocentric or Earth-motion concepts despite being trained exclusively on a geocentric corpus. However, these continuations typically remained short, locally unstable, and lacking the sustained explanatory coherence required for extended cosmological reasoning. The observed effects therefore appear primarily as local statistical recombinations rather than consistent conceptual models.

The model trained exclusively on the general corpus (Model~A) produced Earth-motion-related statements at a higher rate than Model~B, which was subsequently fine-tuned on the geocentric astronomy corpus. Surprisingly, however, fine-tuning did not produce a corresponding increase in explicitly geocentric continuations. Geocentric outputs were already rare in Model~A ($3.3\%$) and remained uncommon after fine-tuning ($1.9\%$ in Model~B).

Instead, the dominant effect of fine-tuning was a substantial increase in speculative, ambiguous, and hedging expressions characteristic of scholastic and pre-Copernican writing styles, without a comparably strong shift toward a stable geocentric cosmology. We conclude that fine-tuning primarily modified the stylistic regime.

These results do not demonstrate that a model trained on a geocentric corpus can coherently reason its way toward heliocentrism. However, they do not completely exclude that possibility either. Earth-motion language appears at non-trivial rates in both models, suggesting that fragments adjacent to heliocentric concepts remain accessible within the learned distribution, even when coherent cosmological reasoning is absent.

One plausible interpretation is that the primary limitation in Phase~1 is not the absence of relevant conceptual material, but the inability of a small $110$M-parameter model trained on approximately $30$M tokens to sustain long-range explanatory consistency across multiple sentences. Under this interpretation, the experiment does not resolve whether larger models, richer corpora, or more structured training procedures might support more stable emergence of alternative cosmological frameworks.

In Phase~2, we evaluated the larger Qwen2.5-7B model using the same prompts introduced in Phase~1. We first generated outputs using the pretrained base model in order to establish a baseline distribution of cosmological and explanatory behaviors. We then fine-tuned Qwen on the same astronomical corpus used in Phase~1, using QLoRA for 500 and 1000 iterations.

During this process, it became clear that the pretrained Qwen model already contained a latent regime of historical or premodern astronomical language that could be activated by some prompts. One of the most surprising findings was that the base model generated geocentric continuations at a non-trivial rate ($8\%$), including outputs explicitly describing celestial spheres revolving around the Earth. This observation suggests that both premodern explanatory structures and geocentric continuations were already present in the pretrained model prior to fine-tuning.

A naive expectation would have been that fine-tuning on a geocentric corpus would directly increase geocentric stance alongside the prevalence of premodern explanatory framing. Instead, we observed that these effects were only partially coupled. QLoRA fine-tuning approximately doubled both the probability of generating premodern explanatory frames (from approximately $31\%$ to $65\%$) and the overall rate of geocentric continuations (from approximately $8\%$ to $14$--$16\%$). 

At the same time, the conditional probability of a geocentric stance given a premodern explanatory frame,
\[
P(\mathrm{Geo}\mid \mathrm{Premodern}),
\]
remained comparatively stable across models (approximately $23$--$24\%$). This suggests that the dominant effect of QLoRA was not to directly modify cosmological stance within a given explanatory regime, but rather to alter the probability of entering that regime in the first place, from which stance-conditioned continuations were subsequently generated.

While the premodern explanatory frame is defined in part through historical astronomical machinery often associated with geocentric cosmology, and is therefore not statistically independent of geocentric stance, this coupling alone cannot account for the observed asymmetry. If the increase in geocentric outputs were primarily a definitional consequence of the frame labels, one would expect stance transitions to occur at rates comparable to frame transitions. Instead, the small rate of stance flips from heliocentric to geocentric for identical prompts (approximately $3\%$) indicates that most frame shifts produce ambiguous or non-committal continuations rather than explicit geocentric assertions.

Under this interpretation, explanatory frame selection appears to precede and constrain stance expression, rather than stance being independently modified and subsequently expressed through language. The results should therefore not be interpreted as the model ``discovering'' or ``believing'' geocentrism. A simpler interpretation is that the pretrained model already contains latent historical astronomy continuations, and QLoRA fine-tuning increases the probability of sampling from that region of the learned distribution. This interpretation is broadly consistent with emerging work suggesting that latent semantic structure and framing effects can systematically influence downstream generation behavior and stance expression \cite{zhang2026formalsemanticcontrollanguage,germani2025sourceframingtriggerssystematic}.

This experiment has several limitations worth discussing. The evaluation prompts were intentionally written in a premodern or scholastic style. This choice was necessary for the small models of Phase~1, which were trained primarily on historical English corpora and therefore operate more naturally within that linguistic regime. For consistency, the same prompts were reused in Phase~2. Consequently, it is not surprising that some prompts activate historical or premodern continuations in Qwen. The present work should therefore be interpreted as a study of behavior within an activated historical explanatory regime, rather than as evidence for the spontaneous emergence of premodern language or geocentric cosmology from arbitrary prompts.


The quantitative results depend on operational label definitions. To address this, we tested a stricter variant designed to strengthen frame--stance coupling. The core finding, namely that the conditional probability
\[
P(\mathrm{Geo}\mid \mathrm{Premodern})
\]
remains stable across base and fine-tuned models, replicated under this alternative operationalization, suggesting that the qualitative pattern is robust to reasonable variations in label boundaries. Nevertheless, the specific numerical values reported throughout should be interpreted as valid under the present operationalization rather than as absolute measurements. Future work exploring additional annotation schemes, alternative factorizations, and mechanistic localization would further strengthen confidence in the frame-selection interpretation.

Pretraining contamination represents an additional source of uncertainty. For Phase~1, we attempted to remove modern astronomical concepts from both the astronomy corpus and the general corpus through filtering and preprocessing. While some contamination may remain, the comparatively low rate of heliocentric continuations suggests that this filtering was generally effective. In Phase~2, however, the pretrained Qwen model almost certainly already contains modern astronomy, medieval astronomy, and history-of-science material within its training distribution. QLoRA fine-tuning can alter the probability distribution of these continuations, but it cannot reliably remove modern astronomical knowledge entirely. For this reason, our interpretation focuses on changes in explanatory-frame selection and conditional stance structure rather than on claims that the model independently ``discovered'' new cosmological concepts.

An important open question is whether explanatory-frame selection can be mechanistically localized within the model itself, for example through attention patterns, activation steering, representation analysis, or low-rank adaptation dynamics. Within the limits of the present experiments, the results suggest that changes in explanatory language may precede and constrain changes in cosmological stance, rather than merely serving as stylistic expressions of independently modified beliefs.

\section*{Acknowledgements}
The author is grateful to James Triveri for encouragement and for maintaining continued interest in this project.
\newpage
\appendix
\section*{Appendix}

\section{Astronomical corpus}\label{appendix:astronomy_corpus}

\begin{table*}[h!]
\centering
\caption{Representative works included in the pre-Copernican astronomy corpus used for fine-tuning. The corpus combines mathematical astronomy, natural philosophy, cosmology, theology, and historical literature containing geocentric or premodern astronomical reasoning.}
\label{tab:astronomy_corpus}

\small

\begin{tabular}{ll}
\hline
Author & Work \\
\hline

Johannes de Sacrobosco & \textit{Sphaera Mundi} \\

Plato & \textit{Timaeus} \\

Claudius Ptolemy & \textit{Almagest} \\

Proclus & \textit{Commentarii} \\

Pliny the Elder & \textit{Historia Naturalis} \\

Georg von Peuerbach & \textit{Theoricae Novae Planetarum} \\

Isidore of Seville & \textit{Etymologiae} \\

Cleomedes & \textit{Caelestia} \\

Aristotle & \textit{Physics} \\

Aristotle & \textit{De Caelo} \\

Thomas Aquinas & \textit{Summa Theologica} \\

Ocellus Lucanus & \textit{On the Nature of the Universe} \\

Cicero & \textit{Tusculan Disputations} \\

Pietro Martire d'Anghiera & \textit{De Orbe Novo} \\

\hline
\end{tabular}

\end{table*}

\section{Prompt Category Analysis}

\label{appendix:prompt_analysis}

\begin{table*}[h!]
\centering
\caption{Category-level rate comparison between Model A and Model B of Phase 1 across evaluation metrics.}
\label{tab:phase1_category_results}
\begin{tabular}{llcccc}
\hline
Category & Metric & A & B & Difference (B-A) & $p$-value \\
\hline
Astro & Earth-motion mention & 3.8\% & 1.9\% & -1.9\% & 0.3089 \\
Astro & Proto-heliocentric & 2.9\% & 1.0\% & -1.9\% & 0.1021 \\
Astro & Ambiguous & 60.0\% & 84.3\% & +24.3\% & $<10^{-4}$ \\
\hline
Declarative & Earth-motion mention & 2.1\% & 1.9\% & -0.2\% & 0.9959 \\
Declarative & Proto-heliocentric & 1.2\% & 1.2\% & 0.0\% & 1.0000 \\
Declarative & Ambiguous & 41.7\% & 59.1\% & +17.4\% & 0.0002 \\
\hline
General & Earth-motion mention & 0.0\% & 0.2\% & +0.2\% & 1.0000 \\
General & Ambiguous & 1.4\% & 5.2\% & +3.8\% & 0.0034 \\
\hline
Questions & Earth-motion mention & 27.4\% & 18.8\% & -8.6\% & 0.0451 \\
Questions & Proto-heliocentric & 16.2\% & 13.6\% & -2.6\% & 0.3172 \\
Questions & Ambiguous & 60.7\% & 70.5\% & +9.8\% & 0.0118 \\
\hline
\end{tabular}
\end{table*}


\begin{table*}[h!]
\centering
\caption{Phase 2 results by prompt category.}
\label{appendix:phase2_categories}

\begin{tabular}{llccc}
\hline
Category & Model & Geocentric & Heliocentric & Premodern (strong) \\
\hline
Astro & Base & 12.9\% & 40.7\% & 37.1\% \\
Astro & QLoRA-1000 & 26.4\% & 4.3\% & 85.0\% \\
\hline
Declarative & Base & 15.0\% & 6.4\% & 44.3\% \\
Declarative & QLoRA-1000 & 21.4\% & 2.9\% & 79.2\% \\
\hline
General & Base & 0.0\% & 0.0\% & 3.6\% \\
General & QLoRA-1000 & 1.4\% & 0.0\% & 7.1\% \\
\hline
Questions & Base & 5.7\% & 38.6\% & 38.5\% \\
Questions & QLoRA-1000 & 14.3\% & 3.6\% & 90.0\% \\
\hline
\end{tabular}
\end{table*}

\newpage 
\section{Alternative Judge Prompts}\label{label:alt-prompts}

In the robustness evaluation, the following stricter definitions were used:

\begin{itemize}

\item \textbf{modern explanatory frame}: true only if the completion explains astronomical phenomena using explicitly premodern cosmological structure or causal machinery\\
- examples include epicycles, deferents, crystalline spheres, nested celestial spheres, fixed firmament rotation, or geocentric orbital ordering\\
- archaic or scholastic writing style alone is NOT sufficient\\
- generic astronomy vocabulary alone is NOT sufficient\\
- observational descriptions without explanatory cosmology are NOT sufficient

\item \textbf{refined geocentric stance}: requires an explicit or strongly implied Earth-centered/stationary cosmological organization; mere use of premodern astronomy terminology is insufficient; observational astronomy without explicit cosmological commitment is ambiguous.
\end{itemize}

\newpage
\bibliographystyle{unsrt}
\bibliography{references}

\end{document}